\documentclass{article} %

\pdfoutput=1
\PassOptionsToPackage{numbers, compress}{natbib}

\usepackage[preprint]{nips_2018}

\usepackage{hyperref}
\usepackage{url}

\usepackage[utf8]{inputenc} %
\usepackage[T1]{fontenc}    %
\usepackage{hyperref}       %
\usepackage{url}            %
\usepackage{booktabs}       %
\usepackage{amsfonts}       %
\usepackage{amssymb}		%
\usepackage{nicefrac}       %
\usepackage{microtype}      %

\usepackage{amsmath}
\usepackage{amsfonts}
\usepackage{bbm}
\usepackage{dsfont}			%
\usepackage[english]{babel}

\usepackage{graphicx}
\usepackage{caption}
\usepackage{subcaption}

\usepackage{adjustbox}
\usepackage{appendix}

\usepackage{enumitem}
\setlist{nosep}

\usepackage{color}
\usepackage[ruled,vlined,linesnumbered]{algorithm2e}

\DeclareMathOperator{\defd}{:=}
\DeclareMathOperator*{\argmax}{arg\,max}

\title{Multi-Preference Actor Critic}

\author{Ishan Durugkar \thanks{ while at Microsoft Research} \\
Department of Computer Science\\
University of Texas at Austin\\
Austin, TX 78712, USA \\
\texttt{ishand@cs.utexas.edu} \\
\And
Matthew Hausknecht, Adith Swaminathan \& Patrick MacAlpine \\
Microsoft Research \\
Redmond, WA \\
\texttt{\{matthew.hausknecht, adswamin,
Patrick.MacAlpine\}@microsoft.com} \\
}

\begin{document}
\maketitle

\begin{abstract}
Policy gradient algorithms typically combine discounted future rewards with an estimated value function, to compute the direction and magnitude of parameter updates.
However, for most Reinforcement Learning tasks, humans can provide additional insight to constrain the policy learning.
We introduce a general method to incorporate multiple different feedback channels into a single policy gradient loss.
In our formulation, the Multi-Preference Actor Critic (M-PAC), these different types of feedback are implemented as constraints on the policy. We use a Lagrangian relaxation to satisfy these constraints using gradient descent while learning a policy that maximizes rewards. Experiments in Atari and Pendulum verify that constraints are being respected and can accelerate the learning process.
\end{abstract}

\section{Introduction}

We examine how to incorporate human preferences into policy gradient reinforcement learning algorithms to achieve higher performance in fewer environment interactions.
Many existing papers have studied how human preferences can be incorporated into reinforcement learning: Human expert demonstrations, one of the more direct expressions of human preference, have been incorporated through a behavior-cloning pre-training phase or by mixing demonstrations with episodic experiences during updates~\citep{hester2017deep,nair2017overcoming,vecerik2017leveraging}.
When an expert is available to provide on-policy feedback, methods such as Dagger~\citep{ross2011dagger} and Aggrevate~\citep{ross2014aggrevate} query the expert to gain access to on-policy demonstrations, reducing the problem of covariate shift. 
Inverse Reinforcement Learning (IRL) attempts to use demonstrations to infer the demonstrator's reward function~\citep{abbeel2004inverseRL,ziebart2008entropy,ho2016gail,finn2016inverse}.
IRL methods are particularly useful in tasks in which there are no explicit rewards and only expert demonstrations are available.

In certain domains, it is difficult for humans to provide direct demonstrations.
Therefore a number of alternate ways of specifying human preferences have been explored:
The TAMER framework \citep{knox09TAMER} uses human feedback as an estimate of the value function.
\cite{macglashan2017coach} show that positive and negative feedback signals provided by humans during the course of an episode can be used to learn advantages over actions.
\cite{christiano2017preferences} show that it is possible to learn complex behavior in environments using only a reward function inferred from asking humans to repeatedly choose between two potential policies.

Other human preferences are encoded as a part of the agent's loss function.
For example, maximum-entropy reinforcement learning~\citep{ziebart2008entropy,haarnoja2017energy} reflects the intuition that a policy should exhibit as much randomness as possible while maximizing rewards.
This preference for greater entropy is expressed as a regularization term applied to the agent's objective function.
Similarly, trust region methods~\citep{schulman2015trust,schulman2017proximal} enforce the human preference that the learning agent's policy should not change drastically between updates. %

Motivated by the variety of human preferences and feedback modalities, we construct a unifying architecture for learning from diverse human preferences.
Multi-Preference Actor Critic (M-PAC) uses a single-actor network paired with multiple critic networks, where each source of preference feedback is encoded by a different critic.
We formulate a constrained optimization problem in which each critic represents a soft constraint applied to the actor's policy, enforcing the corresponding human preference.
This formulation allows us to use a Langrangian relaxation to automatically and dynamically learn the relative weighting of each preference.
For example, in the early stages of the learning process, the agent may place strong emphasis on the critic encouraging similarity to human demonstrations, but later in the learning process, switch emphasis to %
ensure safe policy updates.   

We conduct experiments combining four types of human preferences: entropy regularization, safe policy updates, behavior cloning, and GAIL. 
Our experiments demonstrate that incorporating these preferences as critics in a constrained optimization framework allows faster learning and higher eventual performance. 
Furthermore, using this framework it is possible to incorporate other forms of human feedback in a straightforward manner.

\section{Background}
In this paper, we look at the setting of a Markov
Decision Process $\left<\mathcal{S}, \mathcal{A}, r, \mathcal{P}, \mu, \gamma\right>$ defined by a set of states $\mathcal{S}$, a set of actions $\mathcal{A}$, a reward function $r(s, a, s')$ and transition probabilities $\mathcal{P}(s, a, s') = \Pr(s_{t+1}=s' | s_t=s, a_t=a)$, where $s, s' \in \mathcal{S}$ and $a \in \mathcal{A}$.
$\mu$ is the initial state distribution of the MDP and $0 < \gamma < 1$ is the discount factor.

A policy $\pi(a|s)$ maps states to a probability distribution over actions.
The discounted value of starting in a particular state at time $t$ and then following policy $\pi$ is given by
\begin{align}
V_{\pi}(s) &\defd \mathbb{E}_{\pi}\left[ \sum_{t=0}^{\infty} \gamma^t r(s_t, a_t, s_{t+1}) %
| s_0 = s \right]
\end{align}

The advantage of taking an action $a_t$ in state $s_t$ can be considered as the additional value that the agent would get if it took action $a_t$ and then followed policy $\pi$ from state $s_{t+1}$ over just following policy $\pi$ in state $s_t$.
\begin{align}
	A_{\pi}(s_t, a_t) &\defd r(s_t, a_t, s_{t+1}) + \gamma V_{\pi}(s_{t+1}) - V_{\pi}(s_{t})
\end{align}

The aim of training a reinforcement learning agent is to find a policy that can maximize the agent's value over all states.
\begin{align}\label{eqn:pol_search}
	\pi^* &\defd \argmax_{\pi} \mathbb{E}_{s \sim \mu} V_{\pi}(s)
\end{align}

This search can be done using gradient descent by minimizing the loss $    L = \mathbb{E}_{\pi} - A_{\pi}(s, a)$.
Since $A_{\pi}(s, a)$ cannot be directly optimized in closed form, we typically use the policy gradient trick whereby the policy gradient is $\nabla_{\theta} L = - \mathbb{E}_{\pi_{\theta}} A_{\pi_{\theta}}(s, a)  \log \pi_{\theta}(a|s) $.
In practice, the Advantage Actor Critic method also adds an entropy term ($H$) to the loss, to prevent early convergence to a sub-optimal policy.
\begin{align}
    L = \mathbb{E}_{\pi_{\theta}} \left[- A_{\pi_{\theta}}(s, a) - \beta H(\pi_{\theta}(a|s, \theta)\right]
\end{align}

\section{Multi-Preference Actor Critic} \label{sec:Framework}

Consider $K$ possible preferences.
We define preference $c_k(\pi)$ as a function mapping inputs $\pi$ to $\mathbb{R}^+$.
This preference metric measures the amount by which the preference is violated, and is $0$ if the preference is perfectly satisfied.
As a concrete example, consider a preference on agent behavior expressed through demonstrations $(s_1, a_1, s_2, a_2, \dots)$. We can do behavior cloning to learn a policy $\pi_{bc}$ using this demonstration data. A preference $c_k(\pi)$ can then simply be the KL-divergence between $\pi$ and $\pi_{bc}$: $c_k(\pi) = \mathbb{E}_{\pi} \log \pi(s, a) -\log \pi_{bc}(s, a)$. All the preferences we study in this paper can be viewed as expectations over samples drawn from $\pi$. As shorthand, we express these preferences as $c_k(\pi)=\mathbb{E}_{\pi} d_k(\pi, s, a)$; in the example above, $d_k(\pi, s, a)=\log \pi(s,a) - \log \pi_{bc}(s,a)$.

When we consider incorporating the above preferences into our policy search, observe that each $c_k$ can be naturally interpreted as a critic in actor-critic architectures. So, one possibility is to add the preferences as additional costs incurred by the policy.
The different preferences can then be folded into a single reward function by weighting each cost with a hyper-parameter.

\begin{align}\label{eqn:all_pref}
	L = \mathbb{E}_{\pi} \left[ - A_{\pi}(s, a) + \sum_k \lambda_k d_k(\pi, s, a)) \right]
\end{align}

This is the approach we see in \cite{kang2018policy, gao2018reinforcement, hester2017deep, nair2017overcoming}.
It can be difficult to find the right hyperparameter values to weigh these preferences against each other.
Moreover, the relative usefulness of individual preferences might change as the agent's proficiency increases.

A technique to incorporate varied preferences into the policy learning procedure that can weigh preferences in a principled manner is required.
Consider instead a policy search procedure that is done only in the space of policies that satisfy the preferences.
This leads to a constrained formulation of Equation \ref{eqn:pol_search}.

Consider again the preference metric $c_k(\pi)$.
We can specify how much the policy can stray from this preference by setting a threshold $l_k$.
The search for the optimal policy can then be written as

\begin{align}
	\pi^* &\defd \argmax_{\pi} \mathbb{E}_{s \sim \mu} V_{\pi}(s) \\
    & \text{s.t.   } \forall k \left[ \mathbb{E}_{s \sim \mu}  \sum_a d_k(\pi, s, a) \right] \leq l_k 
\end{align}

If our preferences are sufficiently diverse, the set of policies that satisfies the above constraints will be much smaller than the set of all policies we were searching over when we had only the environmental returns to guide us.

We can now turn to Lagrangian relaxation of these constraints so that they are no longer hard constraints and furthermore, we can use policy gradient to find a feasible policy.
If the agent policy is a function with parameters $\theta$, Lagrangian relaxation with parameters $\lambda_k \in \mathbb{R}^+$ on the constraints leads to the following saddle point problem.

\begin{align}\label{eqn:MPAC}
	\min_{\theta} \max_{\lambda} \mathbb{E}_{\pi_\theta} \left[ - A_{\pi_\theta}(s, a) + \sum_k \lambda_k \left(d_k(\pi_\theta,s,a) - l_k\right) \right].
\end{align}

When we use policy gradients or any other stochastic gradient method to optimize this saddle point formulation, the $\lambda_k$ weight is increased if the preference is violated beyond our threshold.
It decreases to $0$ if the preference metric stays within that threshold.
Policy gradient simultaneously updates $\theta$ to minimize the joint objective.
Equation \ref{eqn:MPAC} looks remarkably like Equation \ref{eqn:all_pref}, but also offers a principled way to adjust the weighting on the preferences.

\section{Examples of Preferences} \label{sec:Preferences}

We now consider how the preferences that we discussed in Section \ref{sec:Related} can be incorporated in the M-PAC framework we set up in Section \ref{sec:Framework}.
To incorporate preferences we convert a preference into the form of a function that maps $\pi, s, a$ to $\mathbb{R}^+$.
We show below that this conversion is fairly straightforward for all the preferences we have considered.

\subsubsection{Entropy}

A high entropy policy is a common preference that is usually enforced as a regularization by means of a surrogate loss \citep{mnih2016asynchronous, haarnoja2017energy}.
Such an entropy loss ensures exploration and prevents premature convergence to a locally optimal policy.

However, the gradient of this loss acts on the policy even when it is nowhere close to being deterministic.
We can instead present a high entropy policy as a preference, and define the entropy preference $d_{entropy}(\pi,s)$ as

\begin{align}
    d_{entropy}(\pi_{\theta},s) &= KL(\pi_{\theta}(s)||q) \\
    \forall a \in \mathcal{A}: q(a) &= \frac{1}{\|\mathcal{A}\|}
\end{align}

\subsubsection{Conservative Updates}
Conservative Policy Iteration \citep{kakade2002approximately}, TRPO \citep{schulman2015trust} and PPO \citep{schulman2017proximal} have shown ensuring updates to the policy do not cause a large divergence from the current policy helps to ensure that the policy performance does not change too drastically.

Instead of constraining the update, however, we introduce a preference of staying close to a previous policy.
Let $\theta'$ be the parameters of this older policy.
The conservative policy preference is given by

\begin{align}
    d_{conserve}(\pi_{\theta}, s) &= KL(\pi_{\theta}(s)||\pi_{\theta'}(s))
\end{align}

We update the previous policy by moving its parameters slowly closer to the training policy.
\begin{align}
    \theta' \leftarrow \eta \theta + (1 - \eta) \theta'
\end{align}

\subsubsection{Reference Policy}
Suppose we have access to a reference policy $\pi_{ref}$ that we believe does fairly well.
This reference policy might have been learned by Behavior Cloning, from observations \citep{torabi2018behavioral} or the output of a scripted agent.
We can add a preference that our agent's policy tries to follow this policy.
Similar to the ones above, this preference can also be modeled as a divergence measure.

\begin{align}
    d_{reference}(\pi_{\theta}, s) &= KL(\pi_{\theta}(s)||\pi_{ref}(s))
\end{align}

In our experiments we learn this reference policy by behavior cloning from demonstrations. 
These demonstrations are typically provided by a human or a trained agent.

\subsubsection{Inverse RL}
When the preference we want to model can be expressed as a policy, modeling it as the divergence of the learned policy from this preferred policy is natural.
However, preferences can be more expressive than that, and can be expressed as a cost-to-go function.

Expert demonstrations can also be used to infer a reward function, which can then be used to compute the advantages used in policy gradient Reinforcement Learning \citep{ho2016gail, abbeel2004inverseRL}.
As an example of the expressiveness of our preference measure, we consider Inverse RL via GAIL~\citep{ho2016gail}.
GAIL uses transitions from an expert policy in the environment along with an agent's transitions from its current policy to generate an adversarial reward function $r_{gail}(s, a)$.

This reward can then be used as any reward in reinforcement learning to learn a policy that mimics the expert.
We learn a value function $V_{gail}(s)$ and use it to calculate the advantage $A_{gail}(s, a)$ of taking an action.
The GAIL preference is then defined as
\begin{align}
    d_{gail}(\pi_{\theta}, s, a) = - \log \pi(a|s, \theta) A_{gail}(s, a)
\end{align}

\section{Experiments}
Our experiments seek to answer the following questions:
\begin{itemize}[leftmargin=2em]
    \item Does M-PAC respect the constraints imposed by the preferences while maximizing reward?
    \item How much impact do different preferences have on the learning process, and at what times in learning are certain preferences preferred over others?
    \item Do domain specific preferences such as GAIL and Reference Policies accelerate the learning?
\end{itemize}

We instantiated M-PAC using Advantage Actor Critic (A2C) as the base policy gradient learning algorithm upon which the preferences in Section \ref{sec:Preferences} were incorporated. 
To show that M-PAC is capable of incorporating various types of preferences, Section \ref{sec:ablations} compares M-PAC and A2C on the Pendulum domain.
An ablation analysis is also performed to analyze the effect of the different preferences in this domain.

Next, Section \ref{sec:pacman_demo} examines the effects of incorporating human demonstrations using the Behavioral Cloning and GAIL preferences on the Ms. Pac-man domain.
Finally, Section \ref{sec:atari_results} evaluates M-PAC on multiple Atari games without demonstrations using the entropy and conservative update preferences.
Experimental details and hyper-parameters are included in Appendix \ref{sec:exp_details}.

\subsection{Pendulum}
\label{sec:ablations}

\begin{figure}[t]
\hspace{-3em}
\begin{subfigure}{.6\textwidth}
  \centering
  \includegraphics[width=\linewidth]{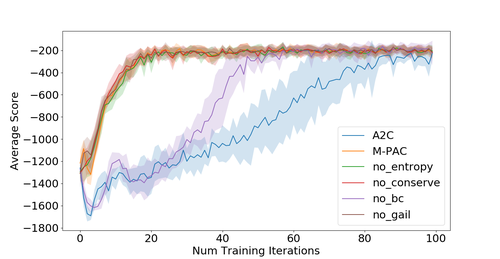}
  \caption{Demonstrations}
  \label{fig:pend_demo}
\end{subfigure}%
\hspace{-2em}
\begin{subfigure}{.6\textwidth}
  \centering
  \includegraphics[width=\linewidth]{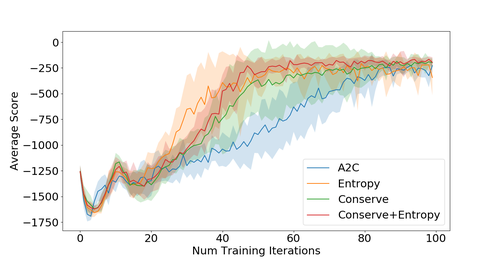}
  \caption{No Demonstrations}
  \label{fig:pend_nodemo}
\end{subfigure}
\caption{\textbf{Pendulum ablation studies} remove one preference of M-PAC at a time and compare to A2C to assess the relative importance of each preference. In Figure \ref{fig:pend_demo}, we utilize demonstrations provided by a policy learned by A2C to learn the reference policy and GAIL reward. In Figure \ref{fig:pend_nodemo} we compare the performance when no demonstrations are available. We see that both entropy and conservative update preferences help to guide learning compared to vanilla A2C.}
\label{fig:ablations}
\end{figure}

To understand how each preference contributes to overall performance, we focus on the Pendulum environment from OpenAI gym.
Refer to Figure \ref{fig:ablations} for the results.

We save demonstrations by running a pre-trained policy trained using A2C that gets an average score of $-195$.
We run this policy in the environment to get 10000 demonstrations.
These demonstrations are then used for behavior cloning and for learning the GAIL reward.
Both the A2C and M-PAC policies are learned through a multi-layer perceptron with 2 layers of 512 units each.
We evaluate the agent every 1000 iterations by evaluating its score on 10 episodes.
Figures are plotted by taking an average of 10 independent runs with separate seeds.

From Figure \ref{fig:pend_demo}, we can see that with enough data, both GAIL and Behavior Cloning can provide enough guidance for the agent to explore and reach the optimal behavior faster than A2C.
Learning from a reference policy directly is very beneficial initially, while GAIL provides a more indirect exploration signal due to its adversarial learning procedure.

To compare the benefits of the preferences for conservative updates and a high entropy policy, we consider Figure \ref{fig:pend_nodemo}.
Both these preferences individually and in tandem help the policy learn faster than A2C.
This is interesting because A2C includes entropy regularization already.

\begin{figure}[h]
\centering
\includegraphics[width=0.8\linewidth]{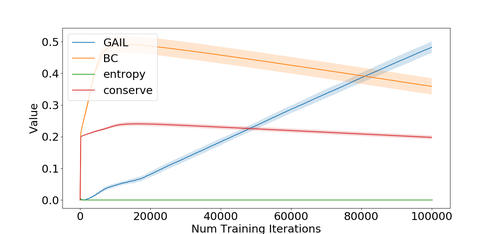}
\caption{\textbf{Change of $\lambda$ values over time} for different preferences in Pendulum. Behavior cloning is heavily used near the beginning of learning while GAIL is increasingly employed during later stages of policy refinement. Conservative updates are used throughout, while entropy isn't needed.}
\label{fig:lambda}
\end{figure}

Another important aspect of M-PAC is the $\lambda$ parameters and how they change with respect to violated constraints.
Figure \ref{fig:lambda} compares the $\lambda$ parameters for the different preferences considered.
Here we see that the $\lambda$ associated with behavior cloned reference policy preference (BC) and conservative update preference (conserve) shoots up to a value where it can control these values according to the threshold we set.
$\lambda$ associated with the GAIL advantages keeps climbing steadily as the advantages keep providing signal to the policy, and is eventually weighted more than the behavior cloned reference policy.
In all this while, the policy entropy does not stray far enough away from a high entropy reference to cause its $\lambda$ to increase from $0$.

\subsection{Using Human Demonstrations}
\label{sec:pacman_demo}

One of the most widely used human feedback mechanisms is learning from demonstrations.
This experiment seeks to measure the effectiveness of human demonstrations incorporated into M-PAC's policy gradient update using the Behavioral Cloning and GAIL preferences.

We collected six episodes of demonstration from a single non-expert play-tester for the Ms. Pac-Man game. M-PAC uses these demonstrations in two ways: First they are used to estimate the human's policy via supervised learning, and constrain the learned policy to stay close to the human's policy. Second, the demonstrations are used to train the GAIL preference's discriminator to discern between M-PAC and human trajectories.

\begin{figure}[h!]
\captionsetup{justification=centering}
\centering
\includegraphics[width=\linewidth]{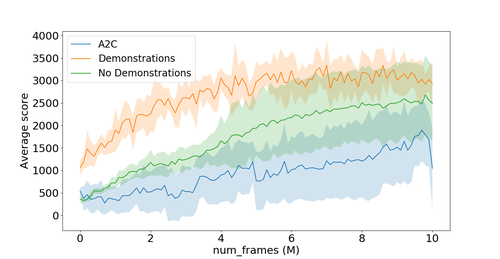}
\caption{\textbf{Benefit of demonstrations}: M-PAC performance on Ms. Pac-Man with and without access to human demonstrations.}
\label{fig:pacman_demo}
\end{figure}

Figure \ref{fig:pacman_demo} compares the performance of M-PAC using the high entropy and conservative update preferences (no demonstrations) to M-PAC with demonstrations additionally using the Behavior-cloning and GAIL preferences.
Comparisons are done across 10 independent runs of each method.
As expected, M-PAC derives clear benefits from preferences that use demonstrations in the form of accelerated policy learning.

\subsection{Without Demonstrations}
\label{sec:atari_results}
In order to assess the general applicability of M-PAC, we perform a rough parameter sweep on seven well-known Atari games with no demonstrations.
For these experiments M-PAC uses the entropy and conservative updates preferences which do not require additional human feedback.
We choose to compare M-PAC to A2C and PPO baselines because the entropy preference can be considered a replacement for the entropy regularization used in A2C, while the conservative update preference is similar in effect to the trust-region updates in PPO.

Each algorithm is trained over 10 million steps (40 million in game frames) across 16 parallel training environments, and PPO results are considered as reported in \citep{schulman2017proximal}.
They are each run 10 times independently with different seeds.
Preliminary results in Table \ref{table:results} and learning curves in Appendix \ref{sec:learning_curves} show that M-PAC is competitive with PPO and A2C across a variety of games.

\begin{table}
\centering
 \begin{tabular}{||c | c c c||} %
 \hline
 Game & A2C & PPO & M-PAC \\ [0.5ex]  %
 \hline\hline
 MsPacman & 1686.1 & 2096.5 & \textbf{2495.22} \\
 \hline
 BeamRider & 3031.7 & 1590.0 & \textbf{3157.36} \\
 \hline
 Breakout & 303.0 & 274.8 & \textbf{326.675} \\
 \hline
 Pong & 19.7 & \textbf{20.7} & 20.41 \\ %
 \hline
 Seaquest & \textbf{1714.3} & 1204.5 & 908.33 \\ %
 \hline
 SpaceInvaders & 744.5 & \textbf{942.5} & 600 \\ %
 \hline
 Qbert & 5879.25 & \textbf{14293.3} & 3769.37 \\ [0.5ex]  %
 \hline
\end{tabular}
\vspace{1em}
\caption{{\small \textbf{Comparison on ALE} Using only conservative updates and entropy regularization, M-PAC outperforms A2C and PPO on three games, ties on one, and performs worse on three.}}
\label{table:results}
\end{table}

\section{Discussion}

Motivated by the diversity and effectiveness of human feedback, we present an algorithm for incorporating multiple human preferences to guide an agent's search for a high-performing policy.
To implement this idea we formalize preferences as constraints on the policy, which can be softened by Lagrangian relaxation to allow policy gradient.
This Lagrangian relaxation also allows the policy to stray away from the preferred policies if its value function indicates better returns.

We examined four different preferences in the M-PAC framework and experimentally evaluated it on Pendulum and Atari environments, and validated that access to even non-expert human demonstrations helps accelerate the search for high-quality policies.

One possible downside of the M-PAC framework is the inability to find an optimal policy if the preferences don't contain the optimal policy in or near their constraint set.
This might happen if the preferences contradict each other, or if the preferences are not good indicators of optimal behaviors.
Alternatively, if the constraints are set too tight, the lambda values might grow prohibitively and take over as the main component of the loss signal. We occasionally observed this phenomenon when using human demonstrations on Atari games.

In future work we seek to consider environment-specific preferences, which can provide very strong signals for learning a proficient policy. Humans quickly understand the interactions in games and could encode this knowledge as preferences. For example, to avoid specific entities such as ghosts in Ms. Pac-Man.
Beyond the exploration in this paper, it is our hope that M-PAC provides an extensible framework for incorporating novel forms of human feedback that may arise in future work.

\clearpage
\bibliographystyle{plainnat}
\bibliography{ishand_bibliography}

\clearpage
\appendix
\appendixpage

\section{ALE Learning Curves}
\label{sec:learning_curves}
\begin{figure}[h]
\begin{subfigure}{.5\textwidth}
  \centering
  \includegraphics[width=\linewidth]{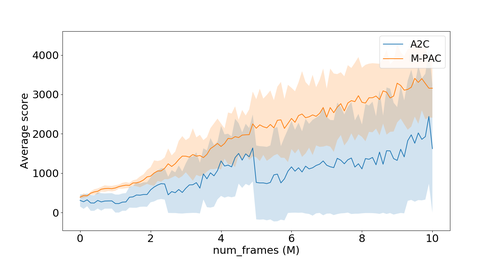}
  \caption{Beamrider}
  \label{fig:beamrider_compare}
\end{subfigure}
\begin{subfigure}{.5\textwidth}
  \centering
  \includegraphics[width=\linewidth]{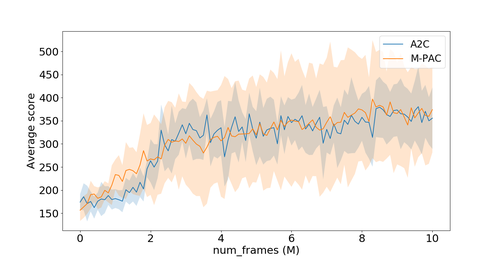}
  \caption{Space Invaders}
  \label{fig:spaceinvaders}
\end{subfigure}
\\
\begin{subfigure}{.5\textwidth}
  \centering
  \includegraphics[width=\linewidth]{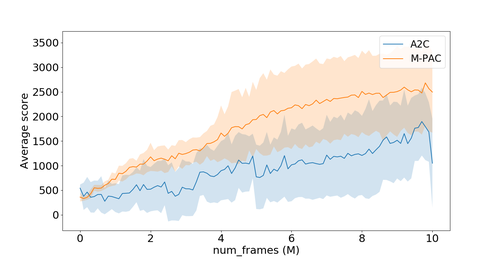}
  \caption{Ms. Pacman}
  \label{fig:pacman}
\end{subfigure}
\begin{subfigure}{.5\textwidth}
  \centering
  \includegraphics[width=\linewidth]{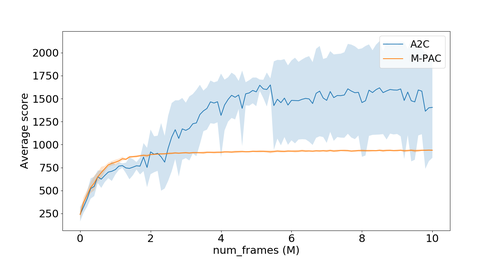}
  \caption{Seaquest}
  \label{fig:seaquest_compare}
\end{subfigure}
\\
\begin{subfigure}{.5\textwidth}
  \centering
  \includegraphics[width=\linewidth]{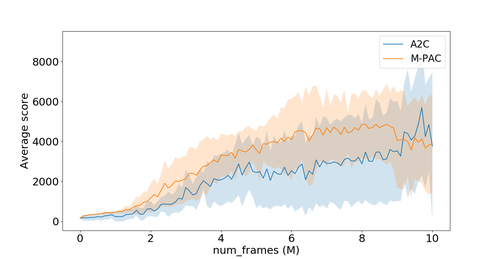}
  \caption{Qbert}
  \label{fig:qbert}
\end{subfigure}%
\begin{subfigure}{.5\textwidth}
  \centering
  \includegraphics[width=\linewidth]{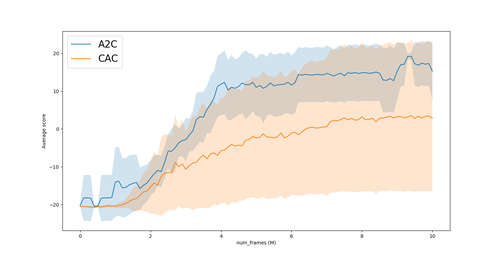}
  \caption{Pong}
  \label{fig:pong_compare}
\end{subfigure}
\\
\begin{subfigure}{.5\textwidth}
  \centering
  \includegraphics[width=\linewidth]{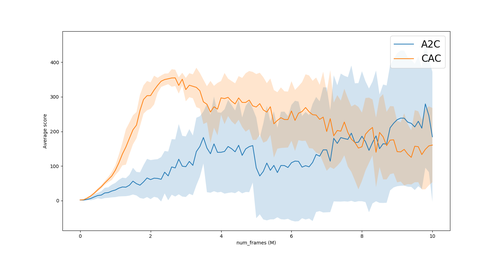}
  \caption{Breakout}
  \label{fig:breakout_compare}
\end{subfigure}

\caption{Runs on ALE comparing M-PAC with A2C. M-PAC uses the High Entropy and Conservative update preferences.}
\label{fig:ale}
\end{figure}

\section{Experiment Details}
\label{sec:exp_details}

Atari experiments are conducted in several well-studied games in the Atari Learning Environment \citep{bellemare2013arcade} using the OpenAI Gym \citep{brockman2016openai} environment wrapper.
Experiments are conducted using the Tensorflow library version 1.10.
Diversity was ensured by seeding each run with a different seed.
Every method was run 10 times with seeds set from $0$ to $9$.
This seed was used to set the random seed for tensorflow, numpy and the test environment using the Gym wrapper.
The seed was also used as a multiplier in setting a seed for the parallel training environments.
After each epoch, the current policy is evaluated on the test environment.

\subsection{Acrobot}

Network: Multi-layer perceptron with 2 hidden layers, 512 units each, ReLU activations\\
Training and Reference policy for Behavior Cloning are separate with identical architectures. Behavior Cloning network includes dropout layers with dropout rate $20\%$ \\
Policy Learning rate: $10^{-4}$ \\
Policy Optimizer: ADAM\\
A2C entropy factor: 0.1\\
Lagrangian Learning rate: $10^{-4}$ \\
Policy Optimizer: Gradient Descent\\
Entropy threshold: $2.$ \\
Conservative Update Threshold: $0.03$ \\
GAIL Threshold: $0.1$ \\
Behavior Cloning Threshold: $0.1$ \\
Number of parallel envs: 8 \\
Epoch: 1000 steps \\
Number of epochs: 100\\
evaluation episodes: 10 \\
All the lagrangian parameters are initialized at $0$.\\
Experiments were conducted on a CPU

\subsection{ALE}

Network: A2C network\\
\begin{itemize}
    \item Convolutional layer 1: 32 filters, size 8, stride 4
    \item Convolutional layer 2: 64 filters, size 4, stride 2
    \item Convolutional layer 1: 64 filters, size 3, stride 1
    \item Fully connected layer: 512 units
\end{itemize}
Training and Behavior Cloned Reference policy share the features of this fully connected hidden layer.
Behavior cloning additionally applies dropout of $20\%$ on top of these features. \\
Policy Learning rate: $7 \times 10^{-4}$ \\
Policy Optimizer: ADAM\\
A2C entropy factor: 0.01\\
Lagrangian Learning rate: $10^{-4}$ \\
Policy Optimizer: Gradient Descent\\
Entropy threshold: $1.5$ \\
Conservative Update Threshold: $0.03$ \\
GAIL Threshold: $0.1$ \\
Behavior Cloning Threshold: $0.1$ \\
Number of parallel envs: 16 \\
Epoch: 6250 steps \\
Number of epochs: 100\\
evaluation episodes: 20 \\
All the lagrangian parameters are initialized at $0$ \\
Experiments were conducted on a Nvidia K40 GPUs.

\end{document}